# Synthesizing multi-layer perceptron network with ant lion, biogeography-based dragonfly algorithm evolutionary strategy invasive weed and league champion optimization hybrid algorithms in predicting heating load in residential buildings

Hossein Moayedi, Amir Mosavi

Oxford Brookes University; a.mosavi@brookes.ac.uk

**Abstract:** The significance of heating load (HL) accurate approximation is the primary motivation of this research to distinguish the most efficient predictive model among several neural-metaheuristic models. The proposed models are through synthesizing multi-layer perceptron network (MLP) with ant lion optimization (ALO), biogeography-based optimization (BBO), dragonfly algorithm (DA), evolutionary strategy (ES), invasive weed optimization (IWO), and league champion optimization (LCA) hybrid algorithms. Each ensemble is optimized in terms of the operating population. Accordingly, the ALO-MLP, BBO-MLP, DA-MLP, ES-MLP, IWO-MLP, and LCA-MLP presented their best performance for population sizes of 350, 400, 200, 500, 50, and 300, respectively. The comparison was carried out by implementing a ranking system. Based on the obtained overall scores (OSs), the BBO (OS = 36) featured as the most capable optimization technique, followed by ALO (OS = 27) and ES (OS = 20). Due to the efficient performance of these algorithms, the corresponding MLPs can be promising substitutes for traditional methods used for HL analysis.

## 1 Introduction

Energy consumption analysis of buildings is a very significant task, due to the high rate of energy consumed in this sector [1]. Heating, ventilating, and air conditioning (HVAC) [2] is a state-of-the-art system that controls the heating load (HL) and cooling load (CL) in the buildings. Considering the crucial importance of the subject, the approaches, like regression analysis [3, 4] and time series [5], cannot satisfy the accuracy required for estimating these parameters. As well as this, other difficulties like the non-linearity of the problem have driven many scholars to benefit the flexibility of intelligent models. As discussed by many scholars, along with well-known models (e.g., decision-making [6-9]), the artificial intelligence techniques have provided a high capability in the estimation of non-linear and intricate parameters [10-12]. Plenty of scientific efforts (e.g., concerning environmental subjects [13-23], gas consumption modeling [24, 25], sustainabile developments [26], pan evaporation and soil precipitation simulation [26-31], energy-related estimations [32-39], water supply assessment [16, 40-49], computer vision and visual processing [50-57], building and structural design analysis [8, 58-61], behavior of structural components [60, 62-64], measurement techniques [43, 50, 65, 66], climatic-related calculations [64], and analysis that deal with feature selection [64, 67-72]) have been associated with these computational technologies. In artificial neural network (ANN), for example, a mapping process between the input and target parameters is carried out by mimicking the neural-based method established in the human's brain [73-76]. Different structures (and consequently diverse types) of ANNs have been designed for specific objectives (e.g., multi-layer perceptron (MLP) [77-79]). Going into deep processors like ANN, a so-called method "deep learning" emerges which has successfully modeled various phenomena and parameters [8, 80-82]. Diagnostic problem and medical sciences, for instance, are two subjects which have been nicely solved by extreme machine learning approaches [83-86].

Up to now, diverse notions of soft computing techniques (e.g., support vector machine (SVM) and artificial neural network (ANN)) have been effectively used for energy consumption modeling [87-91]. Roy, et al. [92] proposed MARS (multivariate adaptive regression splines) coupled with ELM (extreme learning machine) for predicting the HL and CL. They used the first model to importance analysis of the parameters to feed the second model. Likewise, Sholahudin and Han [93] used an ANN along with the Taguchi method for investigating the effect of the input factors on the HL. The feasibility of a random forest predictive method was investigated by Tsanas and Xifara [94] and Gao, et al. [95] for both HL and CL factors. The latter reference is a comprehensive comparative study that compares the simulation capability of sixteen machine learning models (e.g., elastic net, radial basis function regression). This study also confirmed the high efficiency of alternating model tree and rules decision table models. Chou and Bui [91] proposed the combination of ANN and SVM as a proper model for new designs of energy-conserving buildings. The applicability of the neuro-fuzzy approach (ANFIS) for predicting the HL and CL was explored by Nilashi, et al. [96]. They used expectation-maximization and principal component analysis along with the ANFIS, respectively for clustering objective and removing noise. Referring to obtained values of mean absolute error (MAE) (0.16 and 0.52 for the HL and CL predictions, respectively), they concluded the proposed model is accurate enough for this aim.

Also, studies in different fields have shown that utilizing metaheuristic algorithms is an effective idea for improving the accuracy of typical predictors [97, 98]. For energy-efficient buildings, Moayedi, et al. [99] improved the ANN for forecasting the CL by benefiting from the foraging/social behavior of ants, Harris hawks, and elephant (i.e., the EHO algorithm). The results (e.g., the correlation values over 85 %) showed that the applied algorithms can satisfactorily handle the optimization task. An EHO-based CL predictive formula was also presented. Al-Shammari, et al. [100] used the firefly algorithm to optimize the SVM (parameters) for HL modeling in district heating systems. Their model outperformed genetic programming and ANN. Moayedi, et al. [99] employed a grasshopper optimization algorithm (GOA) and grey wolf optimization (GWO) algorithms for enhancing the HL prediction of ANN. A significant decrease in the MEA calculated for the ANN (from 2.0830 to 1.7373 and 1.6514, respectively by incorporation of the GOA and GWO) means that the algorithms can build a more reliable ANN network compared to the typical back-propagation one. Also, other studies like [26] outline the competency of such algorithms in the same fields. As a visible gap of knowledge, despite the variety of studies that have mainly focused on broadly used metaheuristic techniques [101], there are still some algorithms that need to be evaluated. Therefore, assessing the performance of six novel optimization techniques, namely ant lion optimization (ALO), biogeography-based optimization (BBO), many-objective sizing optimization [102-104], data-driven robust optimization [35, 105], dragonfly algorithm (DA), evolutionary strategy (ES), invasive weed optimization (IWO), and league champion optimization (LCA) is the central core of the present paper.

## 2   Data provision and analysis

Providing a reliable dataset is an essential step in intelligent model implementation. This data is used in two stages. Firstly, the significant share is analyzed by the models to infer the relationship between the intended factors and independent variables. The rests are then used to represent unseen conditions of the problem and the performance of the model for stranger data.

In this article, the used dataset is downloaded from a freely available data repository "http://archive.ics.uci.edu/ml/datasets/Energy+efficiency" based on a study by Tsanas and Xifara [94]. They analyzed 768 residential buildings with different geometries using Ecotect software [106] to obtain the HL and CL as the outputs. They set the information of eight independent factors, namely relative compactness (RC), overall height (OH), surface area (SA), orientation, wall area (WA), glazing area (GA), roof area (RA), and glazing area distribution (GAD). Figure 1 shows the distribution of these factors versus the HL, which is aimed to be predicted in this study. Based on plenty of previous studies [97], a random division process is carried out to specify 538 samples (i.e., 70 % of the whole) and 230 rows (i.e., 30 % of the whole) to the training and testing sets, respectively.

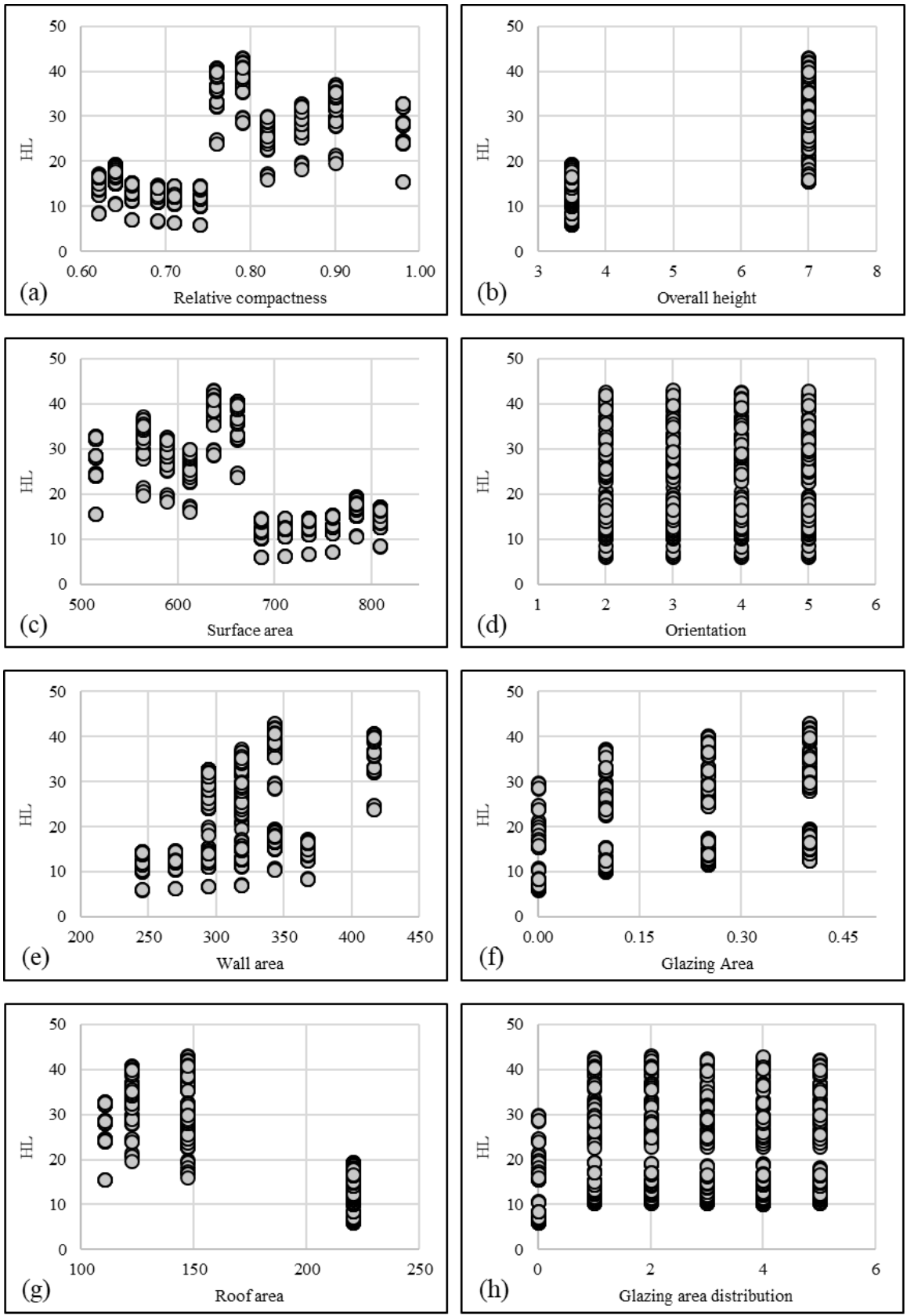

Figure 1: The distribution of the HL versus environmental factors.

## 3 Methodology

The overall methodology used in this study is shown in Figure 2.

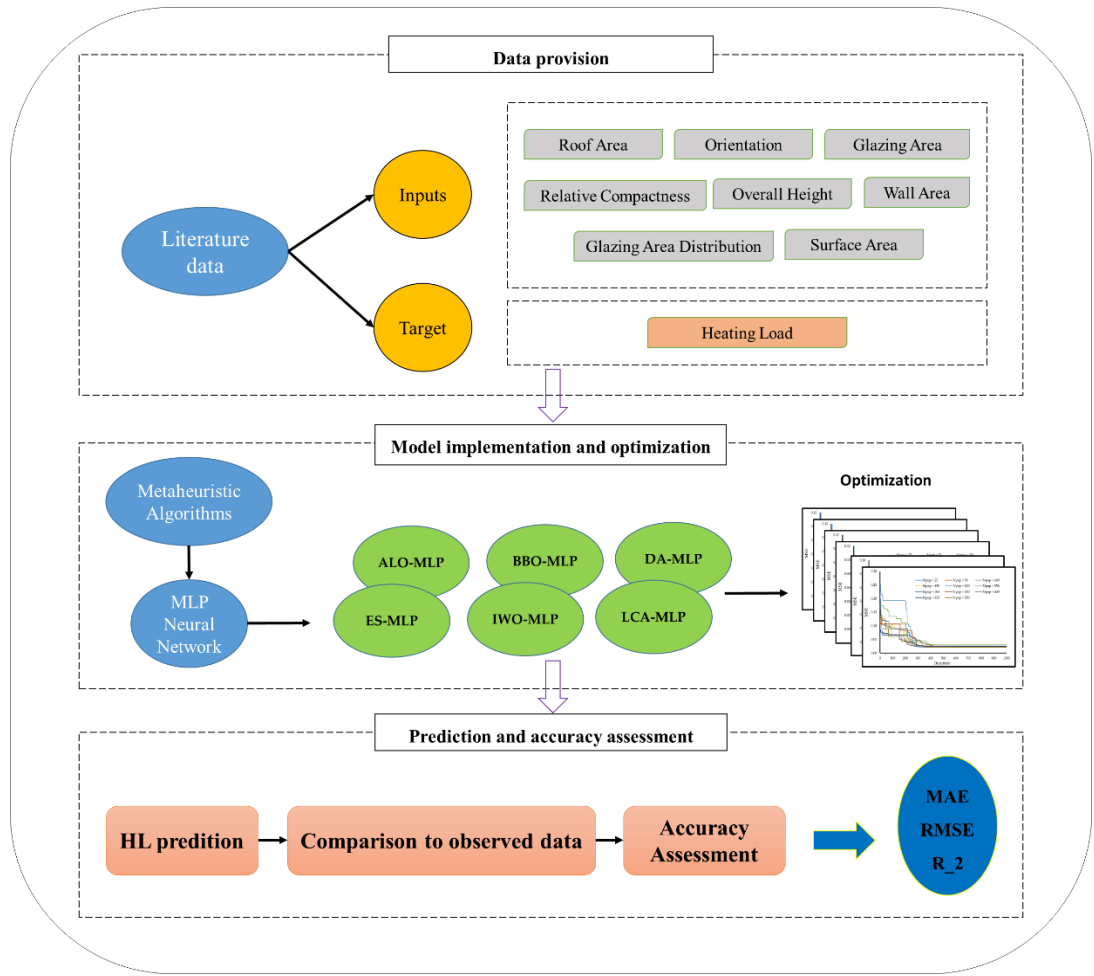

Figure 2: The general path of the study.

### 3.1 Artificial neural network

ANNs are taking part as popular data-mining techniques which are based on the biological mechanism of the neural network [107]. The ANNs are able to deal with highly complicated engineering simulations because of the non-linear analysis option [108, 109]. This approach distinguishes its self by different notions including multi-layer perceptron (MLP) [110], radial basis function [111], and general regression [112]. In this study, an MLP network is selected as the basic method. Figure 3 depicts the MLP general structure predicting $M$ output variables by taking into consideration $L$ input factors. This is proper to note that in an MLP, more than one hidden layer can be sandwiched between two other layers. However, theoretical studies have demonstrated the efficiency of unique hidden layer MLPs for any problem.

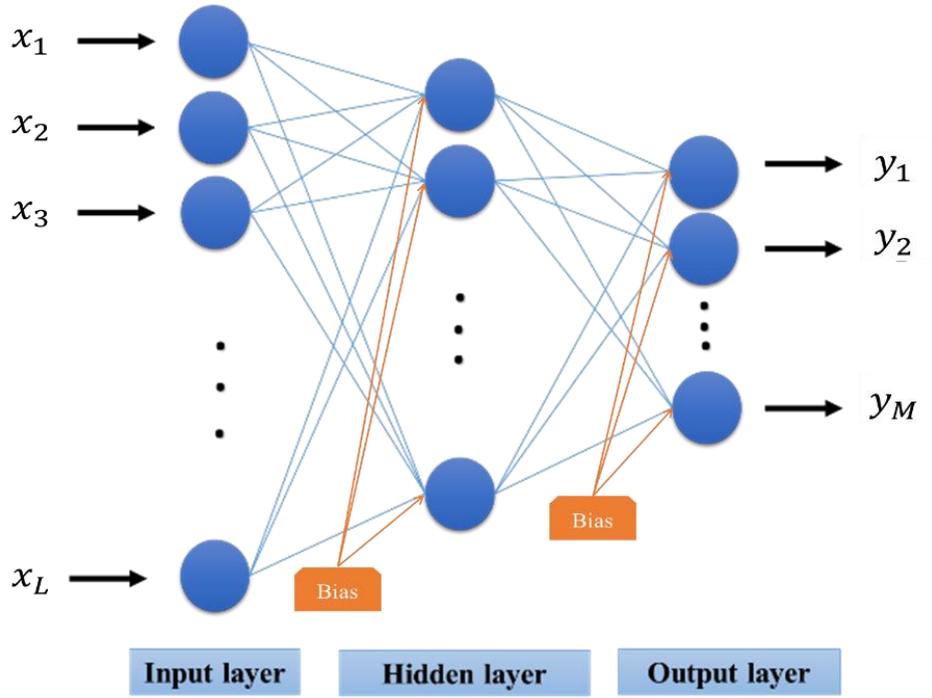

Figure 3: MLP general structure predicting M output variables

The ANNs normally benefit from the training scheme of Levenberg–Marquardt (LM), an approximation to the method of Newton [113] (Equation 1). The LM is known to be quicker and enjoy more power compared to conventional gradient descent technique [114, 115].

$$\Delta \underline{x} = -\left[\nabla^2 V(\underline{x})\right]^{-1} \nabla V(\underline{x}) \qquad (1)$$

in which, $\nabla V(\underline{x})$ and $\nabla^2 V(\underline{x})$ are gradient and the Hessian matrix, respectively. The following equation expresses $V(x)$ as a sum of squares function:

$$V(\underline{x}) = \sum_{i=1}^{N} e_i^2(x) \qquad (2)$$

Next, let $J(x)$ be the Jacobean matrix, then it can be written:

$$\nabla V(\underline{x}) = J(\underline{x}) e(\underline{x})$$
$$\nabla^2 V(\underline{x}) = J^T(\underline{x}) J(\underline{x}) + S(\underline{x}), \qquad (3)$$
$$S(\underline{x}) = \sum_{i=1}^{N} e_i \nabla^2 e_i(\underline{x})$$

Equation 1 can be written as follows when $S(x) \approx 0$:

$$\Delta \underline{x} = \left[J^T(\underline{x}) J(\underline{x})\right]^{-1} J^T(\underline{x}) \underline{e}(\underline{x}) \qquad (4)$$

Lastly, Equation 5 presents the central equation of the LM, based on the Gauss-Newton method.

$$\Delta \underline{x} = \left[J^T(\underline{x}) J(\underline{x}) + \mu I\right]^{-1} J^T(\underline{x}) \underline{e}(\underline{x}) \qquad (5)$$

Remarkably, high and low values of $\mu$ turn this algorithm to steepest descent (with step $1/\mu$) and Gauss-Newton, respectively.

## 3.2 Swarm-based metaheuristic ideas

Optimization algorithms which have been recently very popular for enhancing the performance of predictive models (e.g., the ANNs) are based on swarm functioning of a group of corresponding individuals. They are mostly inspired by nature and seek an optimal global solution for a defined problem by analyzing the relationship between the existing parameters. Coupled with an ANN, these optimizers mean to adjust the biases and weights. This process is better explained in the next section. Here, the overall idea of the intended algorithms is briefly described.

Ant lion optimization Mirjalili [116] is a recently-developed hybrid model that mimics the herding behavior of ant lions. It comprises different stages in which the prey (usual ant) gets trapped and hunted in a hole by a random walk. The capability of the individuals is evaluated by a "roulette wheel selection" function. Biogeography-based optimization is based on two items: (a) the information concerning biogeography and (b) the way different species are distributd. This algorithm was designed by Simon [117] and was used by Mirjalili, et al. [118] to train an MLP network. In the BBO, there are migration and mutation steps and the population is made up of "habits". Note that, these habits are evaluated by two indices called habitat suitability index and suitability index variable. Dragonfly algorithm is another population-based optimization technique proposed by Mirjalili [119]. Based on the Reynolds swarm intelligence, the DA draws on three stages, namely separation, alignment, and cohesion. The name evolutionary strategy implies a stochastic search approach proposed by Schwefel [120]. In the ES, two operators of selection and mutation act during the evolution and adaption stages. The population is produced with offspring variables and the offspring's modality is compared to that of the parents. Inspired by the colonizing behavior of weeds, invasive weed optimization was presented by Mehrabian and Lucas [121]. The optimal solution of this algorithm is the ,most suitable site for the plants to grow and reproduce. The algorithm begins with the initialization and after reproducing, it runs the stages called spatial dispersal and competitive exclusion, and gets stopped after meeting with the termination measures. Last but not least, league champion optimization is suggested by Kashan [122], mimicking sporting competitions in leagues. The LCA tries to find the best-fitted solution to the problem by implementing an artificial league including schedule programming and determining the winner/looser teams. More information about the mentioned algorithms (e.g., mathematical relationships) are details in previous studies (for the ALO [123, 124], BBO [125], DA [126], ES [127], IWO [128], and LCA [129, 130]).

## 3.3 Hybridization process and sensitivity analysis

In order to develop the proposed neural-metaheuristic ensembles, the algorithms should be hybridized with the ANN. To this end, utilizing the provided data, the general equation of an MLP neural network is yielded to the ALO, BBO, DA, ES, IWO, and LCA as the problem function. But before that, it is required to determine the most proper structure (i.e., the number of neurons) of it. As explained previously, the number of neurons in the first and the last layers equals the number of input and output variables, respectively. Hence, it is only the number of hidden neurons that can be varied. Based on a trial-and-error process, it was set to be 5. Therefore, the network architecture is distinguished as $8 \times 5 \times 1$.

Each ensemble was executed within 1000 repetitions, where the mean square error (MSE) was defined to measure the performance error during those (Objective function = MSE). For more reliability of the results, a sensitivity analysis was carried out in this part. Eleven different population sizes, including (25, 50, 100, 150, 200, 250, 300, 350, 400, 450, and 500) were tested for each model and the best-fitted complexity is used to predict the HL in the following. The convergence curves belonging to elite networks of each model are presented in Figure 4. According to these charts, for all algorithms, the error is chiefly reduced within the first half of the iterations. Test best population sizes are determined 350, 400, 200, 500, 50, and 300 for the ALO-MLP, BBO-MLP, DA-MLP, ES-MLP, IWO-MLP, and LCA-MLP, respectively.

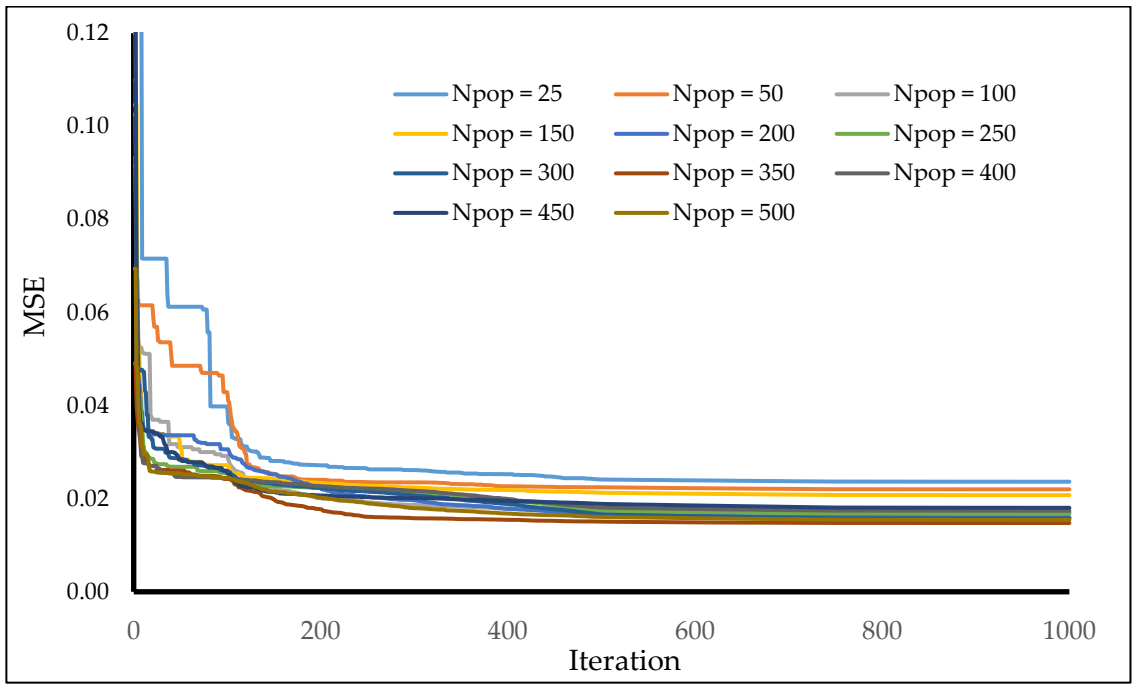

(a)

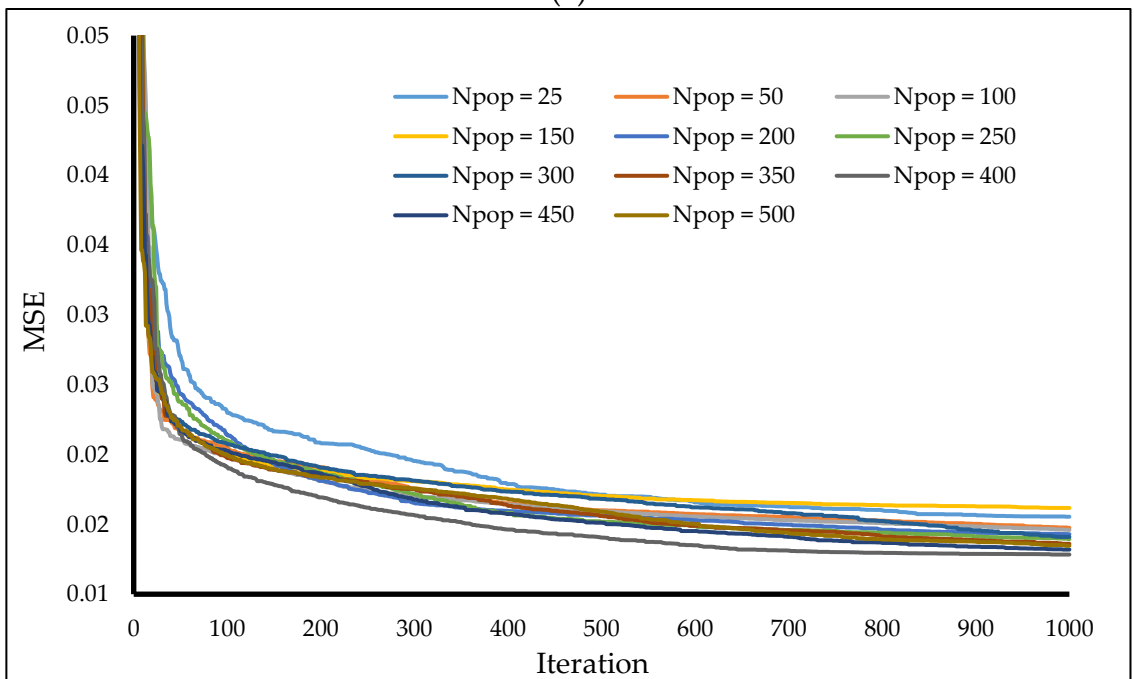

(b)

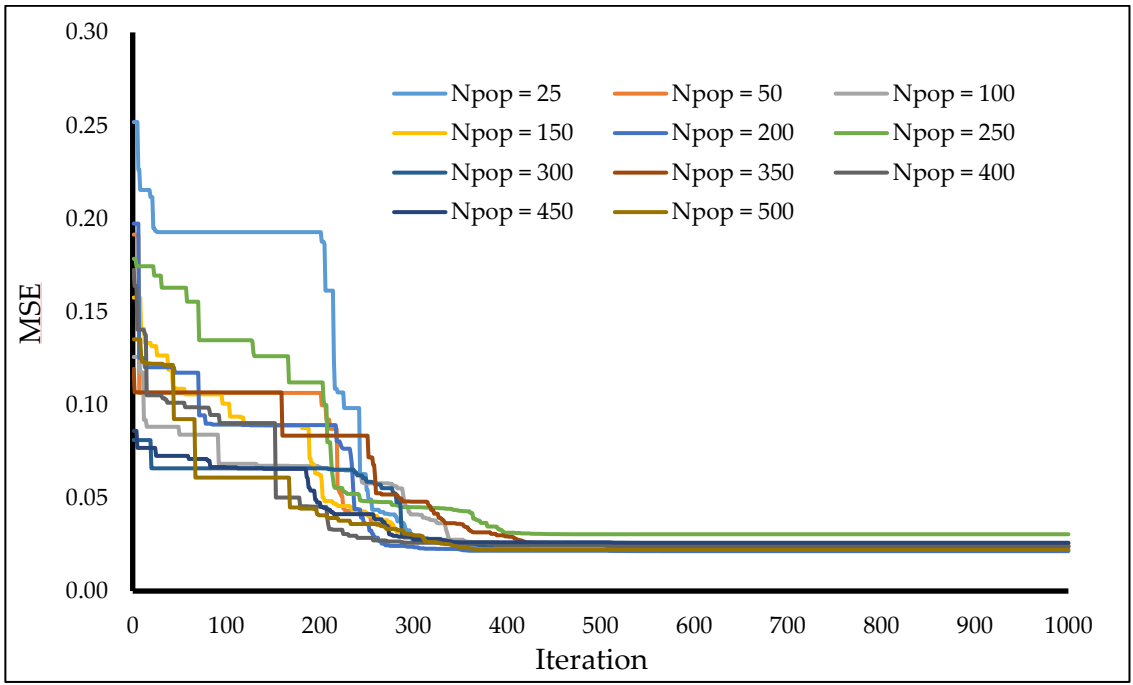

(c)

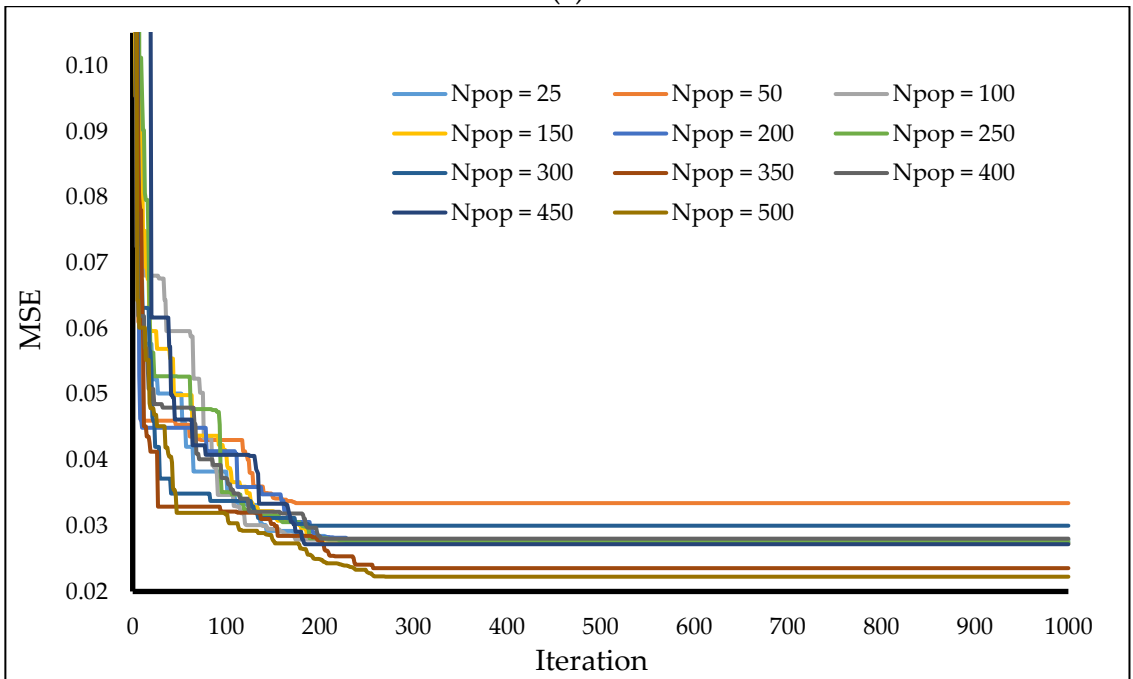

(d)

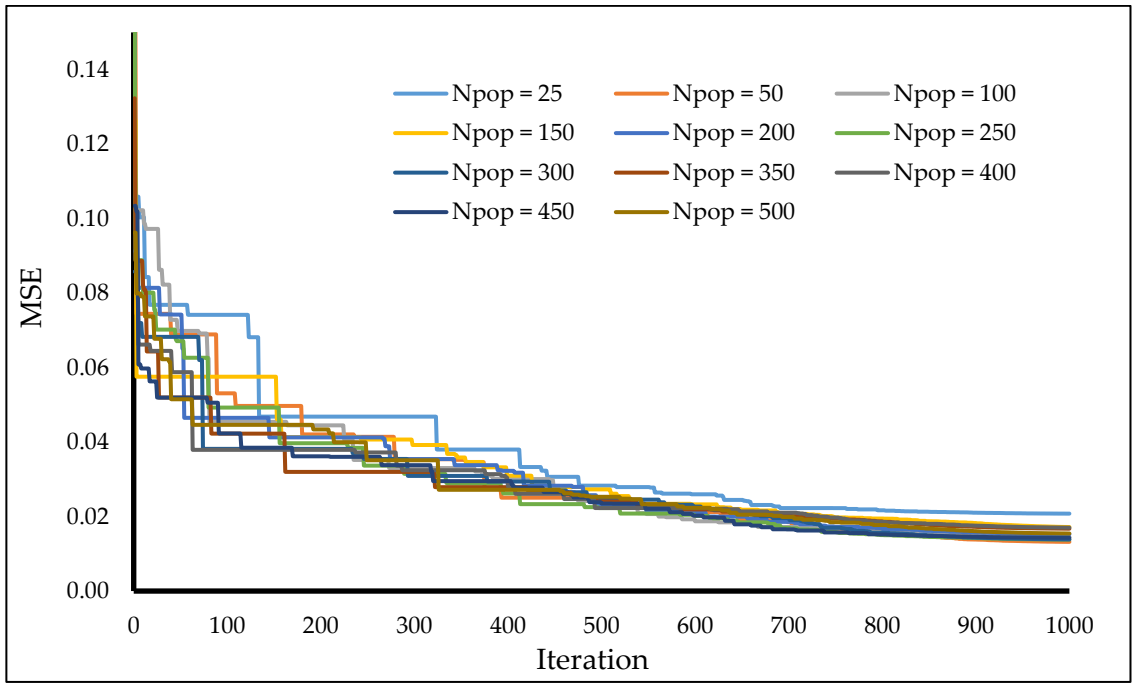

(e)

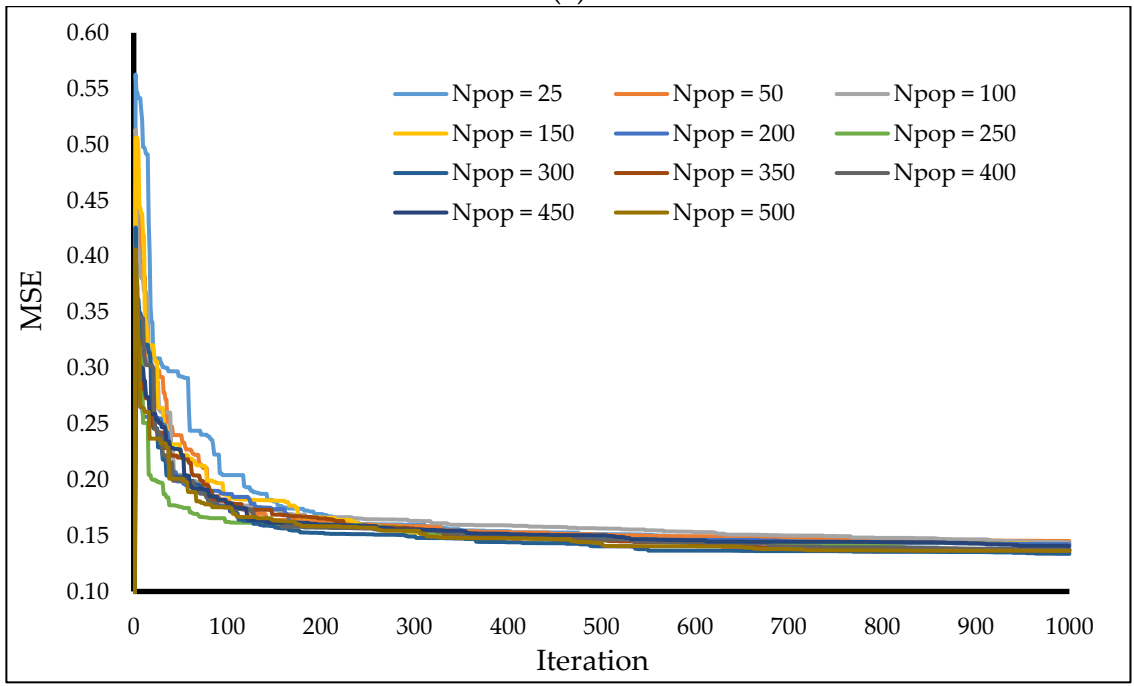

(f)

Figure 4: The sensitivity analysis accomplished for determining the best population size of the (a) ALO-MLP, (b) BBO-MLP, (c) DA-MLP, (d) ES-MLP, (e) IWO-MLP, and (f) LCA-MLP.

## 4 Results and discussion
### 4.1 Statistical accuracy assessment

Three broadly used criteria are applied to measure the prediction accuracy of the implemented models by reporting the error and correlation of the results. For this purpose, MAE (along with the RMSE) and the coefficient of determination ($R^2$) are used. These criteria are applied to the data belonging to the training and testing groups to demonstrate the qualities of learning and prediction, respectively. Assuming $G$ as the total number of samples, $J_{i\ observed}$, and $J_{i\ predicted}$ as the real and forecasted HL values, Equations 6 to 8 formulate the RMSE, MAE, and $R^2$.

$$RMSE = \sqrt{\frac{1}{G}\sum_{i=1}^{G}[(J_{i_{observed}} - J_{i_{predicted}})]^2} \quad (6)$$

$$MAE = \frac{1}{G}\sum_{I=1}^{G}|J_{i_{observed}} - J_{i_{predicted}}| \quad (7)$$

$$R^2 = 1 - \frac{\sum_{i=1}^{G}(J_{i_{predicted}} - J_{i_{observed}})^2}{\sum_{i=1}^{G}(J_{i_{observed}} - \bar{J}_{observed})^2} \quad (8)$$

where $\bar{J}_{observed}$ denotes the mean of $J_{i\ observed}$ values.

## 4.2 Training results

The results of elite structures of each model are evaluated in this part. Figure 5 shows the training results. In this regard, the error (= real HL – forecasted HL) is calculated and marked for all 538 samples. In this phase, the maximum and minimum of the (positive) error values were 0.0136 and 6.4455, 0.0018 and 6.0681, 0.0019 and 9.2773, 0.0248 and 7.3006, 0.0184 and 6.3776, and 0.0715 and 8.4620, respectively for the leaning process of ALO-MLP, BBO-MLP, DA-MLP, ES-MLP, IWO-MLP, and LCA-MLP ensembles.

Referring to the calculated RMSEs (2.6054, 2.5359, 3.4314, 2.7146, 3.2506, and 3.8297), all six models achieved a reliable performance in understanding the non-linear relationship of the HL and eight influential factors. Another evidence that supports this claim is the MAE index (2.0992, 2.0846, 2.9402, 2.0848, 2.8709, and 3.4091). Furthermore, the correlation between the expected and real HLs is higher than 92 % in all models. In details, the values of $R^2$ are 0.9539, 0.9596, 0.9222, 0.9357, 0.9547, and 0.9386.

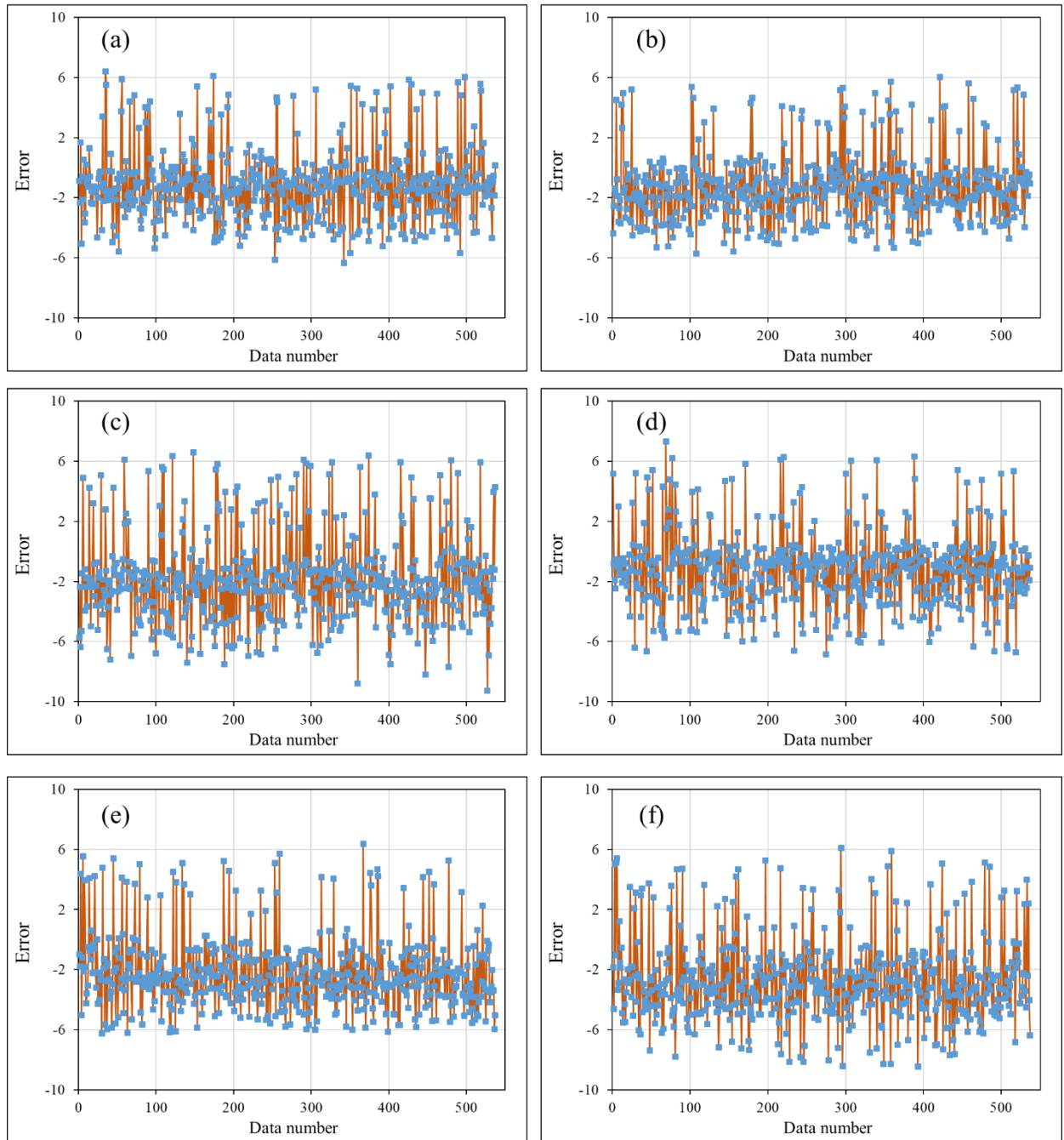

Figure 5: The training errors calculated for the (a) ALO-MLP, (b) BBO-MLP, (c) DA-MLP, (d) ES-MLP, (e) IWO-MLP, and (f) LCA-MLP prediction.

### 4.3 Validation results

The developed models are then applied to the second group of data to assess the generalization capability of them. Figure 6 depicts the correlation between the expected HLs and networks' products. As is seen, all obtained R²s (0.9406, 0.9516, 0.9340, 0.9318, 0.9431, and 0.9400) reflect higher than 93 % accuracy for all models. In this phase, the errors range in [-5.5792, 6.9349], [-5.6311, 6.3000], [-9.3137, 6.8288], [-7.0282, 7.0647], [-6.2505, 5.8823], and [-8.2384, 6.1992].

Considering the computed RMSE (2.7162, 2.4807, 3.3998, 3.0958, 3.3524, and 3.2954) as well as the MAE (2.1865, 1.8284, 2.8713, 2.5072, 2.9702, and 2.7807) error criteria, it can be deduced that the networks' prediction for unseen environmental conditions is in a good level of accuracy. More clearly, the values of mean absolute percentage error were 10.01%, 9.78%, 13.59%, 12.63%, 13.01%, 13.01%, respectively.

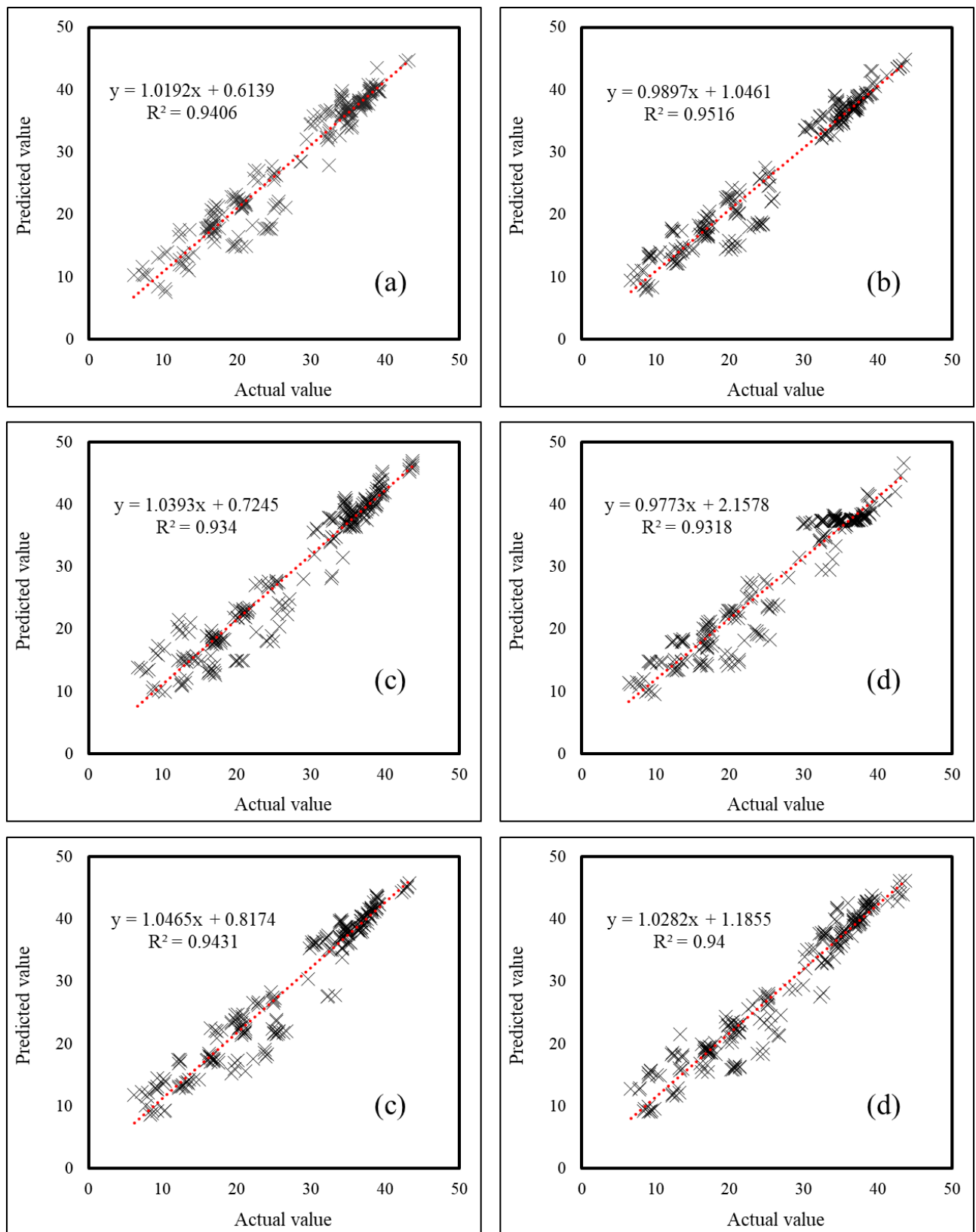

Figure 6: The R² results calculated in the testing phase of the (a) ALO-MLP, (b) BBO-MLP, (c) DA-MLP, (d) ES-MLP, (e) IWO-MLP, and (f) LCA-MLP models.

## 4.4 Score-based comparison and time efficiency

Table 1 summarizes the values of the RMSE, MAE, and $R^2$ obtained for the training and testing phases. In this section, the comparison between the performance of the used predictors is carried out to determine the most reliable one. For this purpose, by taking into consideration all three accuracy criteria, a ranking system is developed. In this way, a score is calculated for each criterion based on the relative performance of the proposed model. The summation of these scores gives an overall score (OS) to rank the models. Table 2 gives the scores assigned to each model.

According to the results, the most significant OS (= 18) is obtained for the BBO-MLP in both training and testing phases. The ALO and ES-based ensembles emerged as the second and third accurate ones, respectively. However, the IWO in the training phase and the LCA in the testing phase gained a similar rank with the ES. Also, it can be seen that the results of the DA-MLP are less consistent than other colleagues.

Table 1: The results of accuracy assessment.

| Ensemble models | Network results | | | | | |
| --- | --- | --- | --- | --- | --- | --- |
| | Training phase | | | Testing phase | | |
| | RMSE | MAE | $R^2$ | RMSE | MAE | $R^2$ |
| **ALO-MLP** | 2.6054 | 2.0992 | 0.9539 | 2.7162 | 2.1865 | 0.9406 |
| **BBO-MLP** | 2.5359 | 2.0846 | 0.9596 | 2.4807 | 1.8284 | 0.9516 |
| **DA-MLP** | 3.4314 | 2.9402 | 0.9222 | 3.3998 | 2.8713 | 0.9340 |
| **ES-MLP** | 2.7146 | 2.0848 | 0.9357 | 3.0958 | 2.5072 | 0.9318 |
| **IWO-MLP** | 3.2506 | 2.8709 | 0.9547 | 3.3524 | 2.9702 | 0.9431 |
| **LCA-MLP** | 3.8297 | 3.4091 | 0.9386 | 3.2954 | 2.7807 | 0.9400 |

Table 2: The executed ranking system.

| Models | Scores | | | | | | | | | |
| --- | --- | --- | --- | --- | --- | --- | --- | --- | --- | --- |
| | Training | | | | | Testing | | | | |
| | RMSE | MAE | $R^2$ | Overall score | Rank | RMSE | MAE | $R^2$ | Overall score | Rank |
| **ALO-MLP** | 5 | 4 | 4 | 13 | 2 | 5 | 5 | 4 | 14 | 2 |
| **BBO-MLP** | 6 | 6 | 6 | 18 | 1 | 6 | 6 | 6 | 18 | 1 |
| **DA-MLP** | 2 | 2 | 1 | 5 | 5 | 1 | 2 | 2 | 5 | 6 |
| **ES-MLP** | 4 | 5 | 2 | 11 | 3 | 4 | 4 | 1 | 9 | 3 |
| **IWO-MLP** | 3 | 3 | 5 | 11 | 3 | 2 | 1 | 5 | 8 | 5 |
| **LCA-MLP** | 1 | 1 | 3 | 5 | 5 | 3 | 3 | 3 | 9 | 3 |

Moreover, Figure 7 illustrates the time required for implementing the used models. This item is also measured for other well-known optimization techniques (including HHO [131], GWO [132], WOA [133], ABC [134], ACO [135], EHO [136], GA [137], ICA [138], PSO [139], and WDO [140]) to be compared with ALO, BBO, DA, ES, IWO, and LCA. This figure indicates that the metaheuristic algorithms used in this study present a good time-efficiency in comparison with other models. Moreover, it was observed that the ABC, HHO, and DA take the largest time for almost all of the population sizes.

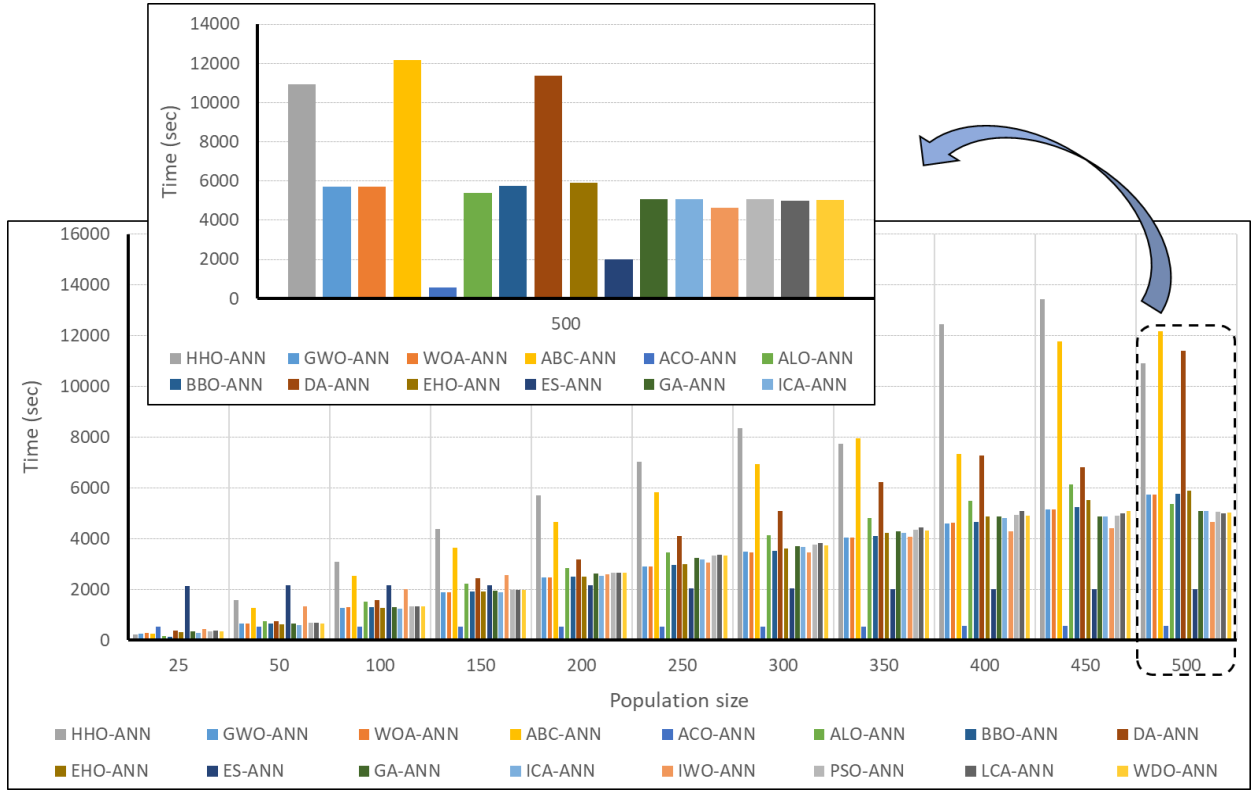

Figure 7: The computation time needed for various hybrid methods.

## 4.5 Presenting the HL predictive equation

In the previous section, it was concluded that the BBO constructs the most reliable neural network. It means that the biases and connecting weights optimized by this technique can analyze and predict the HL more accurately compared to other metaheuristic algorithms. Therefore, the governing relationships in the BBO-MLP ensemble are extracted and presented as the best HL predictive formula (Equation 9). As is seen, there are five parameters ($Z1$, $Z2$, …, $Z5$) in this equation, which need to be calculated by Equation 10. Basically, the response of the neurons in the hidden layer are represented by $Z1$, $Z2$, …, $Z5$. Remarkably, the term *Tansig* is the network activation function, which is expressed by Equation 11.

$$HL_{BBO\text{-}MLP} = 0.9076 \times Z1 + 0.0050 \times Z2 - 0.3986 \times Z3 - 0.4754 \times Z4 - 0.2692 \times Z5 + 0.0283 \quad (9)$$

$$\begin{bmatrix} Z1 \\ Z2 \\ Z3 \\ Z4 \\ Z5 \end{bmatrix} = Tansig \left( \left( \begin{bmatrix} -0.8459 & 0.2944 & -0.7562 & 0.1225 & -0.2456 & 0.3266 & -1.0020 & 0.6090 \\ -0.2863 & 0.4134 & -0.1649 & -0.8857 & 0.8828 & -0.9327 & 0.1703 & 0.4336 \\ 0.7094 & -0.5079 & -0.6916 & 0.6346 & -0.3142 & -0.0794 & -0.4306 & 0.9990 \\ -1.1274 & -0.0470 & -0.1336 & 0.6061 & 0.0406 & 0.3088 & -0.8939 & -0.6135 \\ 0.1514 & 0.2735 & -0.8389 & 0.1982 & -0.6465 & -1.0777 & 0.2336 & 0.6753 \end{bmatrix} \begin{bmatrix} RC \\ SA \\ WA \\ RA \\ OH \\ Orientation \\ GA \\ GAD \end{bmatrix} \right) + \begin{bmatrix} 1.7120 \\ 0.8560 \\ 0.0000 \\ -0.8560 \\ 1.7120 \end{bmatrix} \right) \quad (10)$$

$$Tansig\ (x) = \frac{2}{1 + e^{-2x}} - 1 \qquad (11)$$

## 4.6 Further discussion and future works

Due to the fact that the dataset used in this study is a prepared dataset dedicated to residential buildings, the applicability of the used methods is derived for this types of building. However, there are many researches that have successfully employed machine learning tools for predicting the thermal loads of buildings with other usages like office, commercial, and industrial ones [141]. Hence, utilizing multi-usage datasets for future works can overcome this limitation.

Another idea may be evaluating the accuracy of new generation of hybrid models which can be divided into a) combination of the existing metaheuristic tools with other intelligent models e.g., ANFIS and SVM or b) utilizing more recent optimizers for the existing ANN models. Both ideas are helpful to recognize possibly more efficient predictive methods. Moreover, a practical use of the implemented models is also of interest. In order to evaluate the generalizability of the methods, they can be applied to the information taken from real-world buildings noting that the input parameters considered for predicting the HL should be the same as those used in this study; otherwise, it would be a new development.

## 5 Conclusions

The high competency of optimization techniques in various engineering fields motivated the authors to employ and compar the efficacy of six novel metaheuristic techniques, namely ant lion optimization, biogeography-based optimization, dragonfly algorithm, evolutionary strategy, invasive weed optimization, and league champion optimization in hybridizing the neural network for accurate estimation of the heating load. The proper structure of all seven methods was determined by sensitivity analysis and it was shown that the most appropriate population size could be varied from one algorithm to one another. The smallest and largest populations were 50 and 500 hired by the IWO and ES, respectively. The high rate of accuracy observed for all models indicated that metaheuristic techniques could successfully establish a non-linear ANN-based relationship that predicts the HL from the building characteristics. Comparison based on the used accuracy indices revealed that the BBO, ALO, and ES (with around 94 % correlation of the results) are able to construct more reliable ANNs in comparison with IWO, LCA, and DA. Also, the models enjoy a good time-efficiency relative to some other existing algorithms. However, the authors believe that, due to recent advances in metaheuristic science, further comparative studies may be required for outlining the most efficient predictive method.

**Author Contributions: H.M.** methodology; software validation, writing—original draft preparation, A.M. writing—review and editing, visualization, supervision, project administration. All authors have read and agreed to the published version of the manuscript.

**Funding:** This research was not funded

**Conflicts of Interest:** The authors declare no conflict of interest.

# 6 References

1. IEA, International Energy Agency. Key World Energy Statistics. 2015.
2. McQuiston, F. C.; Parker, J. D.; Spitler, J. D., *Heating, ventilating, and air conditioning: analysis and design*. John Wiley & Sons: 2004.
3. Turhan, C.; Kazanasmaz, T.; Uygun, I. E.; Ekmen, K. E.; Akkurt, G. G. J. E.; Buildings, Comparative study of a building energy performance software (KEP-IYTE-ESS) and ANN-based building heat load estimation. **2014,** 85, 115-125.
4. Catalina, T.; Iordache, V.; Caracaleanu, B., Multiple regression model for fast prediction of the heating energy demand. *Energy and Buildings* **2013,** 57, 302-312.
5. Geysen, D.; De Somer, O.; Johansson, C.; Brage, J.; Vanhoudt, D., Operational thermal load forecasting in district heating networks using machine learning and expert advice. *Energy and Buildings* **2018,** 162, 144-153.
6. Liu, S.; Chan, F. T. S.; Ran, W., Decision making for the selection of cloud vendor: An improved approach under group decision-making with integrated weights and objective/subjective attributes. *Expert Systems with Applications* **2016,** 55, 37-47.
7. Tian, P.; Lu, H.; Feng, W.; Guan, Y.; Xue, Y., Large decrease in streamflow and sediment load of Qinghai–Tibetan Plateau driven by future climate change: A case study in Lhasa River Basin. *CATENA* **2020,** 187, 104340.
8. Sun, Y.; Wang, J.; Wu, J.; Shi, W.; Ji, D.; Wang, X.; Zhao, X., Constraints hindering the development of high-rise modular buildings. *Applied Sciences* **2020,** 10, (20), 7159.
9. Liu, S.; Yu, W.; Chan, F. T. S.; Niu, B., A variable weight-based hybrid approach for multi-attribute group decision making under interval-valued intuitionistic fuzzy sets. *International Journal of Intelligent Systems* **2021,** 36, (2), 1015-1052.
10. Liu, Z.; Shao, J.; Xu, W.; Chen, H.; Zhang, Y., An extreme learning machine approach for slope stability evaluation and prediction. *Natural hazards* **2014,** 73, (2), 787-804.
11. Piotrowski, A. P.; Osuch, M.; Napiorkowski, M. J.; Rowinski, P. M.; Napiorkowski, J. J., Comparing large number of metaheuristics for artificial neural networks training to predict water temperature in a natural river. *Computers & Geosciences* **2014,** 64, 136-151.
12. Liu, J.; Wu, C.; Wu, G.; Wang, X., A novel differential search algorithm and applications for structure design. *Applied Mathematics and Computation* **2015,** 268, 246-269.
13. Feng, S.; Lu, H.; Tian, P.; Xue, Y.; Lu, J.; Tang, M.; Feng, W., Analysis of microplastics in a remote region of the Tibetan Plateau: Implications for natural environmental response to human activities. *Science of The Total Environment* **2020,** 739, 140087.
14. Fu, X.; Fortino, G.; Pace, P.; Aloi, G.; Li, W., Environment-fusion multipath routing protocol for wireless sensor networks. *Information Fusion* **2020,** 53, 4-19.
15. Han, X.; Zhang, D.; Yan, J.; Zhao, S.; Liu, J., Process development of flue gas desulphurization wastewater treatment in coal-fired power plants towards zero liquid discharge: Energetic, economic and environmental analyses. *Journal of Cleaner Production* **2020,** 261, 121144.
16. He, L.; Chen, Y.; Zhao, H.; Tian, P.; Xue, Y.; Chen, L., Game-based analysis of energy-water nexus for identifying environmental impacts during Shale gas operations under stochastic input. *Science of The Total Environment* **2018,** 627, 1585-1601.
17. He, L.; Shen, J.; Zhang, Y., Ecological vulnerability assessment for ecological conservation and environmental management. *Journal of Environmental Management* **2018,** 206, 1115-1125.
18. Liu, J.; Liu, Y.; Wang, X., An environmental assessment model of construction and demolition waste based on system dynamics: a case study in Guangzhou. *Environmental Science and Pollution Research* **2020,** 27, (30), 37237-37259.
19. Wang, Y.; Yuan, Y.; Wang, Q.; Liu, C.; Zhi, Q.; Cao, J., Changes in air quality related to the control of coronavirus in China: Implications for traffic and industrial emissions. *Science of The Total Environment* **2020,** 731, 139133.

20. Liu, L.; Li, J.; Yue, F.; Yan, X.; Wang, F.; Bloszies, S.; Wang, Y., Effects of arbuscular mycorrhizal inoculation and biochar amendment on maize growth, cadmium uptake and soil cadmium speciation in Cd-contaminated soil. *Chemosphere* **2018,** 194, 495-503.
21. Yang, Y.; Liu, J.; Yao, J.; Kou, J.; Li, Z.; Wu, T.; Zhang, K.; Zhang, L.; Sun, H., Adsorption behaviors of shale oil in kerogen slit by molecular simulation. *Chemical Engineering Journal* **2020,** 387, 124054.
22. Liu, J.; Yi, Y.; Wang, X., Exploring factors influencing construction waste reduction: A structural equation modeling approach. *Journal of Cleaner Production* **2020,** 276, 123185.
23. Sun, L.; Li, C.; Zhang, C.; Liang, T.; Zhao, Z., The Strain Transfer Mechanism of Fiber Bragg Grating Sensor for Extra Large Strain Monitoring. *Sensors* **2019,** 19, (8), 1851.
24. Su, Z.; Liu, E.; Xu, Y.; Xie, P.; Shang, C.; Zhu, Q., Flow field and noise characteristics of manifold in natural gas transportation station. *Oil & Gas Science and Technology–Revue d'IFP Energies nouvelles* **2019,** 74, 70.
25. Liu, E.; Lv, L.; Yi, Y.; Xie, P., Research on the Steady Operation Optimization Model of Natural Gas Pipeline Considering the Combined Operation of Air Coolers and Compressors. *IEEE Access* **2019,** 7, 83251-83265.
26. Keshtegar, B.; Heddam, S.; Sebbar, A.; Zhu, S.-P.; Trung, N.-T., SVR-RSM: a hybrid heuristic method for modeling monthly pan evaporation. *Environmental Science and Pollution Research* **2019,** 26, (35), 35807-35826.
27. Ghaemi, A.; Rezaie-Balf, M.; Adamowski, J.; Kisi, O.; Quilty, J., On the applicability of maximum overlap discrete wavelet transform integrated with MARS and M5 model tree for monthly pan evaporation prediction. *Agricultural and Forest Meteorology* **2019,** 278, 107647.
28. Kisi, O.; Heddam, S., Evaporation modelling by heuristic regression approaches using only temperature data. *Hydrological Sciences Journal* **2019,** 64, (6), 653-672.
29. Roy, D. K.; Barzegar, R.; Quilty, J.; Adamowski, J., Using ensembles of adaptive neuro-fuzzy inference system and optimization algorithms to predict reference evapotranspiration in subtropical climatic zones. *Journal of Hydrology* **2020**, 125509.
30. Zhang, B.; Xu, D.; Liu, Y.; Li, F.; Cai, J.; Du, L., Multi-scale evapotranspiration of summer maize and the controlling meteorological factors in north China. *Agricultural and Forest Meteorology* **2016,** 216, 1-12.
31. Chao, L.; Zhang, K.; Li, Z.; Zhu, Y.; Wang, J.; Yu, Z., Geographically weighted regression based methods for merging satellite and gauge precipitation. *Journal of Hydrology* **2018,** 558, 275-289.
32. Chen, Y.; He, L.; Guan, Y.; Lu, H.; Li, J., Life cycle assessment of greenhouse gas emissions and water-energy optimization for shale gas supply chain planning based on multi-level approach: Case study in Barnett, Marcellus, Fayetteville, and Haynesville shales. *Energy Conversion and Management* **2017,** 134, 382-398.
33. He, L.; Chen, Y.; Li, J., A three-level framework for balancing the tradeoffs among the energy, water, and air-emission implications within the life-cycle shale gas supply chains. *Resources, Conservation and Recycling* **2018,** 133, 206-228.
34. Lu, H.; Tian, P.; He, L., Evaluating the global potential of aquifer thermal energy storage and determining the potential worldwide hotspots driven by socio-economic, geo-hydrologic and climatic conditions. *Renewable and Sustainable Energy Reviews* **2019,** 112, 788-796.
35. Wu, C.; Wu, P.; Wang, J.; Jiang, R.; Chen, M.; Wang, X., Critical review of data-driven decision-making in bridge operation and maintenance. *Structure and Infrastructure Engineering* **2020**, 1-24.
36. Zhu, L.; Kong, L.; Zhang, C., Numerical Study on Hysteretic Behaviour of Horizontal-Connection and Energy-Dissipation Structures Developed for Prefabricated Shear Walls. *Applied Sciences* **2020,** 10, (4), 1240.
37. Zhao, X.; Ye, Y.; Ma, J.; Shi, P.; Chen, H., Construction of electric vehicle driving cycle for studying electric vehicle energy consumption and equivalent emissions. *Environmental Science and Pollution Research* **2020**, 1-15.

38. Deng, Y.; Zhang, T.; Sharma, B. K.; Nie, H., Optimization and mechanism studies on cell disruption and phosphorus recovery from microalgae with magnesium modified hydrochar in assisted hydrothermal system. *Science of The Total Environment* **2019,** 646, 1140-1154.
39. Zhang, T.; Wu, X.; Fan, X.; Tsang, D. C. W.; Li, G.; Shen, Y., Corn waste valorization to generate activated hydrochar to recover ammonium nitrogen from compost leachate by hydrothermal assisted pretreatment. *Journal of Environmental Management* **2019,** 236, 108-117.
40. Chen, Y.; Li, J.; Lu, H.; Yan, P., Coupling system dynamics analysis and risk aversion programming for optimizing the mixed noise-driven shale gas-water supply chains. *Journal of Cleaner Production* **2021,** 278, 123209.
41. Cheng, X.; He, L.; Lu, H.; Chen, Y.; Ren, L., Optimal water resources management and system benefit for the Marcellus shale-gas reservoir in Pennsylvania and West Virginia. *Journal of Hydrology* **2016,** 540, 412-422.
42. Li, L.-L.; Liu, Y.-W.; Tseng, M.-L.; Lin, G.-Q.; Ali, M. H., Reducing environmental pollution and fuel consumption using optimization algorithm to develop combined cooling heating and power system operation strategies. *Journal of Cleaner Production* **2020,** 247, 119082.
43. Qian, J.; Feng, S.; Tao, T.; Hu, Y.; Li, Y.; Chen, Q.; Zuo, C., Deep-learning-enabled geometric constraints and phase unwrapping for single-shot absolute 3D shape measurement. *APL Photonics* **2020,** 5, (4), 046105.
44. Quan, Q.; Hao, Z.; Xifeng, H.; Jingchun, L., Research on water temperature prediction based on improved support vector regression. *Neural Computing and Applications* **2020**, 1-10.
45. Yang, M.; Sowmya, A., An Underwater Color Image Quality Evaluation Metric. *IEEE Transactions on Image Processing* **2015,** 24, (12), 6062-6071.
46. Cao, B.; Fan, S.; Zhao, J.; Yang, P.; Muhammad, K.; Tanveer, M., Quantum-enhanced multiobjective large-scale optimization via parallelism. *Swarm and Evolutionary Computation* **2020,** 57, 100697.
47. Qian, J.; Feng, S.; Li, Y.; Tao, T.; Han, J.; Chen, Q.; Zuo, C., Single-shot absolute 3D shape measurement with deep-learning-based color fringe projection profilometry. *Optics Letters* **2020,** 45, (7), 1842-1845.
48. Lyu, Z.; Chai, J.; Xu, Z.; Qin, Y.; Cao, J., A Comprehensive Review on Reasons for Tailings Dam Failures Based on Case History. *Advances in Civil Engineering* **2019,** 2019, 4159306.
49. Feng, W.; Lu, H.; Yao, T.; Yu, Q., Drought characteristics and its elevation dependence in the Qinghai–Tibet plateau during the last half-century. *Scientific Reports* **2020,** 10, (1), 14323.
50. Yan, J.; Pu, W.; Zhou, S.; Liu, H.; Bao, Z., Collaborative detection and power allocation framework for target tracking in multiple radar system. *Information Fusion* **2020,** 55, 173-183.
51. Zenggang, X.; Zhiwen, T.; Xiaowen, C.; Xue-min, Z.; Kaibin, Z.; Conghuan, Y., Research on Image Retrieval Algorithm Based on Combination of Color and Shape Features. *Journal of Signal Processing Systems* **2019**, 1-8.
52. Zhu, Q., Research on Road Traffic Situation Awareness System Based on Image Big Data. *IEEE Intelligent Systems* **2020,** 35, (1), 18-26.
53. Xu, S.; Wang, J.; Shou, W.; Ngo, T.; Sadick, A.-M.; Wang, X., Computer Vision Techniques in Construction: A Critical Review. *Archives of Computational Methods in Engineering* **2020**.
54. Sun, L.; Yang, Z.; Jin, Q.; Yan, W., Effect of Axial Compression Ratio on Seismic Behavior of GFRP Reinforced Concrete Columns. *International Journal of Structural Stability and Dynamics* **2020,** 20, (06), 2040004.
55. Chao, M.; Kai, C.; Zhiwei, Z., Research on tobacco foreign body detection device based on machine vision. *Transactions of the Institute of Measurement and Control* **2020,** 42, (15), 2857-2871.
56. Liu, D.; Wang, S.; Huang, D.; Deng, G.; Zeng, F.; Chen, H., Medical image classification using spatial adjacent histogram based on adaptive local binary patterns. *Computers in biology and medicine* **2016,** 72, 185-200.

57. Abedini, M.; Mutalib, A. A.; Zhang, C.; Mehrmashhadi, J.; Raman, S. N.; Alipour, R.; Momeni, T.; Mussa, M. H., Large deflection behavior effect in reinforced concrete columns exposed to extreme dynamic loads. *Frontiers of Structural and Civil Engineering* **2020,** 14, (2), 532-553.
58. Yang, C.; Gao, F.; Dong, M., Energy Efficiency Modeling of Integrated Energy System in Coastal Areas. *Journal of Coastal Research* **2020,** 103, (SI), 995-1001.
59. Xiong, Z.; Xiao, N.; Xu, F.; Zhang, X.; Xu, Q.; Zhang, K.; Ye, C., An Equivalent Exchange Based Data Forwarding Incentive Scheme for Socially Aware Networks. *Journal of Signal Processing Systems* **2020**.
60. Abedini, M.; Zhang, C., Performance assessment of concrete and steel material models in ls-dyna for enhanced numerical simulation, a state of the art review. *Archives of Computational Methods in Engineering* **2020**, 1-22.
61. Mou, B.; Li, X.; Bai, Y.; Wang, L., Shear behavior of panel zones in steel beam-to-column connections with unequal depth of outer annular stiffener. *Journal of Structural Engineering* **2019,** 145, (2), 04018247.
62. Mou, B.; Zhao, F.; Qiao, Q.; Wang, L.; Li, H.; He, B.; Hao, Z., Flexural behavior of beam to column joints with or without an overlying concrete slab. *Engineering Structures* **2019,** 199, 109616.
63. Gholipour, G.; Zhang, C.; Mousavi, A. A., Numerical analysis of axially loaded RC columns subjected to the combination of impact and blast loads. *Engineering Structures* **2020,** 219, 110924.
64. Zhang, H.; Qu, S.; Li, H.; Luo, J.; Xu, W., A Moving Shadow Elimination Method Based on Fusion of Multi-Feature. *IEEE Access* **2020,** 8, 63971-63982.
65. Yang, W.; Pudasainee, D.; Gupta, R.; Li, W.; Wang, B.; Sun, L., An overview of inorganic particulate matter emission from coal/biomass/MSW combustion: Sampling and measurement, formation, distribution, inorganic composition and influencing factors. *Fuel Processing Technology* **2020**, 106657.
66. Zhang, C.-W.; Ou, J.-P.; Zhang, J.-Q., Parameter optimization and analysis of a vehicle suspension system controlled by magnetorheological fluid dampers. *Structural Control and Health Monitoring* **2006,** 13, (5), 885-896.
67. Yue, H.; Wang, H.; Chen, H.; Cai, K.; Jin, Y., Automatic detection of feather defects using Lie group and fuzzy Fisher criterion for shuttlecock production. *Mechanical Systems and Signal Processing* **2020,** 141, 106690.
68. Zhu, G.; Wang, S.; Sun, L.; Ge, W.; Zhang, X., Output Feedback Adaptive Dynamic Surface Sliding-Mode Control for Quadrotor UAVs with Tracking Error Constraints. *Complexity* **2020,** 2020, 8537198.
69. Xiong, Q.; Zhang, X.; Wang, W.-F.; Gu, Y., A Parallel Algorithm Framework for Feature Extraction of EEG Signals on MPI. *Computational and Mathematical Methods in Medicine* **2020,** 2020, 9812019.
70. Zhang, J.; Liu, B., A review on the recent developments of sequence-based protein feature extraction methods. *Current Bioinformatics* **2019,** 14, (3), 190-199.
71. Zhang, X.; Fan, M.; Wang, D.; Zhou, P.; Tao, D., Top-k Feature Selection Framework Using Robust 0-1 Integer Programming. *IEEE Transactions on Neural Networks and Learning Systems* **2020**, 1-15.
72. Zhao, X.; Li, D.; Yang, B.; Chen, H.; Yang, X.; Yu, C.; Liu, S., A two-stage feature selection method with its application. *Computers & Electrical Engineering* **2015,** 47, 114-125.
73. Cao, B.; Zhao, J.; Lv, Z.; Gu, Y.; Yang, P.; Halgamuge, S. K., Multiobjective Evolution of Fuzzy Rough Neural Network via Distributed Parallelism for Stock Prediction. *IEEE Transactions on Fuzzy Systems* **2020,** 28, (5), 939-952.
74. Shi, K.; Wang, J.; Tang, Y.; Zhong, S., Reliable asynchronous sampled-data filtering of T–S fuzzy uncertain delayed neural networks with stochastic switched topologies. *Fuzzy Sets and Systems* **2020,** 381, 1-25.

75. Shi, K.; Wang, J.; Zhong, S.; Tang, Y.; Cheng, J., Non-fragile memory filtering of T-S fuzzy delayed neural networks based on switched fuzzy sampled-data control. *Fuzzy Sets and Systems* **2020,** 394, 40-64.
76. Yang, S.; Deng, B.; Wang, J.; Li, H.; Lu, M.; Che, Y.; Wei, X.; Loparo, K. A., Scalable Digital Neuromorphic Architecture for Large-Scale Biophysically Meaningful Neural Network With Multi-Compartment Neurons. *IEEE Transactions on Neural Networks and Learning Systems* **2020,** 31, (1), 148-162.
77. Hornik, K.; Stinchcombe, M.; White, H., Multilayer feedforward networks are universal approximators. *Neural networks* **1989,** 2, (5), 359-366.
78. Lv, Z.; Qiao, L., Deep belief network and linear perceptron based cognitive computing for collaborative robots. *Applied Soft Computing* **2020,** 92, 106300.
79. Adeli, H., Neural networks in civil engineering: 1989‐2000. *Computer‐Aided Civil and Infrastructure Engineering* **2001,** 16, (2), 126-142.
80. Xu, M.; Li, T.; Wang, Z.; Deng, X.; Yang, R.; Guan, Z., Reducing Complexity of HEVC: A Deep Learning Approach. *IEEE Transactions on Image Processing* **2018,** 27, (10), 5044-5059.
81. Li, T.; Xu, M.; Zhu, C.; Yang, R.; Wang, Z.; Guan, Z., A Deep Learning Approach for Multi-Frame In-Loop Filter of HEVC. *IEEE Transactions on Image Processing* **2019,** 28, (11), 5663-5678.
82. Qiu, T.; Shi, X.; Wang, J.; Li, Y.; Qu, S.; Cheng, Q.; Cui, T.; Sui, S., Deep Learning: A Rapid and Efficient Route to Automatic Metasurface Design. *Advanced Science* **2019,** 6, (12), 1900128.
83. Chen, H.-L.; Wang, G.; Ma, C.; Cai, Z.-N.; Liu, W.-B.; Wang, S.-J., An efficient hybrid kernel extreme learning machine approach for early diagnosis of Parkinson′s disease. *Neurocomputing* **2016,** 184, 131-144.
84. Hu, L.; Hong, G.; Ma, J.; Wang, X.; Chen, H., An efficient machine learning approach for diagnosis of paraquat-poisoned patients. *Computers in Biology and Medicine* **2015,** 59, 116-124.
85. Wang, S.-J.; Chen, H.-L.; Yan, W.-J.; Chen, Y.-H.; Fu, X., Face recognition and micro-expression recognition based on discriminant tensor subspace analysis plus extreme learning machine. *Neural processing letters* **2014,** 39, (1), 25-43.
86. Xia, J.; Chen, H.; Li, Q.; Zhou, M.; Chen, L.; Cai, Z.; Fang, Y.; Zhou, H., Ultrasound-based differentiation of malignant and benign thyroid Nodules: An extreme learning machine approach. *Computer methods and programs in biomedicine* **2017,** 147, 37-49.
87. Li, Q.; Meng, Q.; Cai, J.; Yoshino, H.; Mochida, A., Applying support vector machine to predict hourly cooling load in the building. *Applied Energy* **2009,** 86, (10), 2249-2256.
88. Urdaneta, S.; andJuan Contreras, E. Z., Fuzzy Model for Estimation of Energy Performance of Residential Buildings. *International Journal of Applied Engineering Research* **2017,** 12, (11), 2766-2771.
89. Fan, C.; Xiao, F.; Zhao, Y., A short-term building cooling load prediction method using deep learning algorithms. *Applied Energy* **2017,** 195, 222-233.
90. Xie, L. J. P. C. S., The heat load prediction model based on BP neural network-markov model. **2017,** 107, 296-300.
91. Chou, J.-S.; Bui, D.-K., Modeling heating and cooling loads by artificial intelligence for energy-efficient building design. *Energy and Buildings* **2014,** 82, 437-446.
92. Roy, S. S.; Roy, R.; Balas, V. E., Estimating heating load in buildings using multivariate adaptive regression splines, extreme learning machine, a hybrid model of MARS and ELM. *Renewable and Sustainable Energy Reviews* **2018,** 82, 4256-4268.
93. Sholahudin, S.; Han, H., Simplified dynamic neural network model to predict heating load of a building using Taguchi method. *Energy* **2016,** 115, 1672-1678.
94. Tsanas, A.; Xifara, A., Accurate quantitative estimation of energy performance of residential buildings using statistical machine learning tools. *Energy and Buildings* **2012,** 49, 560-567.

95. Gao, W.; Alsarraf, J.; Moayedi, H.; Shahsavar, A.; Nguyen, H., Comprehensive preference learning and feature validity for designing energy-efficient residential buildings using machine learning paradigms. *Applied Soft Computing* **2019,** 84, 105748.
96. Nilashi, M.; Dalvi-Esfahani, M.; Ibrahim, O.; Bagherifard, K.; Mardani, A.; Zakuan, N., A soft computing method for the prediction of energy performance of residential buildings. *Measurement* **2017,** 109, 268-280.
97. Moayedi, H.; Hayati, S., Artificial intelligence design charts for predicting friction capacity of driven pile in clay. *Neural Computing and Applications* **2019,** 31, (11), 7429-7445.
98. Fu, X.; Pace, P.; Aloi, G.; Yang, L.; Fortino, G., Topology optimization against cascading failures on wireless sensor networks using a memetic algorithm. *Computer Networks* **2020,** 177, 107327.
99. Moayedi, H.; Nguyen, H.; Foong, L., Nonlinear evolutionary swarm intelligence of grasshopper optimization algorithm and gray wolf optimization for weight adjustment of neural network. *Engineering with Computers.* **2019**.
100. Al-Shammari, E. T.; Keivani, A.; Shamshirband, S.; Mostafaeipour, A.; Yee, L.; Petković, D.; Ch, S., Prediction of heat load in district heating systems by Support Vector Machine with Firefly searching algorithm. *Energy* **2016,** 95, 266-273.
101. Le, L. T.; Nguyen, H.; Dou, J.; Zhou, J., A comparative study of PSO-ANN, GA-ANN, ICA-ANN, and ABC-ANN in estimating the heating load of buildings' energy efficiency for smart city planning. *Applied Sciences* **2019,** 9, (13), 2630.
102. Cao, B.; Dong, W.; Lv, Z.; Gu, Y.; Singh, S.; Kumar, P., Hybrid Microgrid Many-Objective Sizing Optimization With Fuzzy Decision. *IEEE Transactions on Fuzzy Systems* **2020,** 28, (11), 2702-2710.
103. Cao, B.; Wang, X.; Zhang, W.; Song, H.; Lv, Z., A Many-Objective Optimization Model of Industrial Internet of Things Based on Private Blockchain. *IEEE Network* **2020,** 34, (5), 78-83.
104. Cao, B.; Zhao, J.; Yang, P.; Gu, Y.; Muhammad, K.; Rodrigues, J. J. P. C.; Albuquerque, V. H. C. d., Multiobjective 3-D Topology Optimization of Next-Generation Wireless Data Center Network. *IEEE Transactions on Industrial Informatics* **2020,** 16, (5), 3597-3605.
105. Qu, S.; Han, Y.; Wu, Z.; Raza, H., Consensus Modeling with Asymmetric Cost Based on Data-Driven Robust Optimization. *Group Decision and Negotiation* **2020**.
106. Roberts, A.; Marsh, A., ECOTECT: environmental prediction in architectural education. **2001**.
107. McCulloch, W. S.; Pitts, W., A logical calculus of the ideas immanent in nervous activity. *The bulletin of mathematical biophysics* **1943,** 5, (4), 115-133.
108. Seyedashraf, O.; Mehrabi, M.; Akhtari, A. A., Novel approach for dam break flow modeling using computational intelligence. *Journal of Hydrology* **2018,** 559, 1028-1038.
109. Moayedi, H.; Mehrabi, M.; Mosallanezhad, M.; Rashid, A. S. A.; Pradhan, B., Modification of landslide susceptibility mapping using optimized PSO-ANN technique. *Engineering with Computers* **2019,** 35, (3), 967-984.
110. Hornik, K., Approximation capabilities of multilayer feedforward networks. *Neural networks* **1991,** 4, (2), 251-257.
111. Orr, M. J., Introduction to radial basis function networks. Technical Report, Center for Cognitive Science, University of Edinburgh: 1996.
112. Specht, D. F., A general regression neural network. *IEEE transactions on neural networks* **1991,** 2, (6), 568-576.
113. Marquardt, D. W., An Algorithm for Least-Squares Estimation of Nonlinear Parameters. *Journal of the Society for Industrial and Applied Mathematics* **1963,** 11, (2), 431-441.
114. El-Bakry, M. Y., Feed forward neural networks modeling for K-P interactions. *Chaos, Solitons and Fractals* **2003,** 18, (5), 995-1000.
115. Cigizoglu, H. K.; Kişi, Ö., Flow prediction by three back propagation techniques using k-fold partitioning of neural network training data. *Hydrology Research* **2005,** 36, (1), 49-64.

116. Mirjalili, S., The ant lion optimizer. *Advances in Engineering Software* **2015,** 83, 80-98.
117. Simon, D., Biogeography-based optimization. *IEEE transactions on evolutionary computation* **2008,** 12, (6), 702-713.
118. Mirjalili, S.; Mirjalili, S. M.; Lewis, A., Let a biogeography-based optimizer train your multi-layer perceptron. *Information Sciences* **2014,** 269, 188-209.
119. Mirjalili, S., Dragonfly algorithm: a new meta-heuristic optimization technique for solving single-objective, discrete, and multi-objective problems. *Neural Computing and Applications* **2016,** 27, (4), 1053-1073.
120. Schwefel, H.-P. P., *Evolution and optimum seeking: the sixth generation*. John Wiley & Sons, Inc.: 1993.
121. Mehrabian, A. R.; Lucas, C., A novel numerical optimization algorithm inspired from weed colonization. *Ecological informatics* **2006,** 1, (4), 355-366.
122. Kashan, A. H. In *League championship algorithm: a new algorithm for numerical function optimization*, 2009 International Conference of Soft Computing and Pattern Recognition, 2009; IEEE: pp 43-48.
123. Špoljarić, T.; Pavić, I. In *Performance analysis of an ant lion optimizer in tuning generators' excitation controls in multi machine power system*, 2018 41st international convention on information and communication technology, electronics and microelectronics (MIPRO), 2018; IEEE: pp 1040-1045.
124. Kose, U., An ant-lion optimizer-trained artificial neural network system for chaotic electroencephalogram (EEG) prediction. *Applied Sciences* **2018,** 8, (9), 1613.
125. Bhattacharya, A.; Chattopadhyay, P. K., Solving complex economic load dispatch problems using biogeography-based optimization. *Expert Systems with Applications* **2010,** 37, (5), 3605-3615.
126. Vanishree, J.; Ramesh, V., Optimization of size and cost of static var compensator using dragonfly algorithm for voltage profile improvement in power transmission systems. *International Journal of Renewable Energy Research (IJRER)* **2018,** 8, (1), 56-66.
127. Yuan, C.; Moayedi, H., The performance of six neural-evolutionary classification techniques combined with multi-layer perception in two-layered cohesive slope stability analysis and failure recognition. *Engineering with Computers* **2019**, 1-10.
128. Ghasemi, M.; Ghavidel, S.; Akbari, E.; Vahed, A. A., Solving non-linear, non-smooth and non-convex optimal power flow problems using chaotic invasive weed optimization algorithms based on chaos. *Energy* **2014,** 73, 340-353.
129. Kashan, A. H., League Championship Algorithm (LCA): An algorithm for global optimization inspired by sport championships. *Applied Soft Computing* **2014,** 16, 171-200.
130. Kashan, A. H., An efficient algorithm for constrained global optimization and application to mechanical engineering design: League championship algorithm (LCA). *Computer-Aided Design* **2011,** 43, (12), 1769-1792.
131. Heidari, A. A.; Mirjalili, S.; Faris, H.; Aljarah, I.; Mafarja, M.; Chen, H., Harris Hawks optimization: Algorithm and applications. *Future Generation Computer Systems* **2019,** 97, 849-872.
132. Mirjalili, S.; Mirjalili, S. M.; Lewis, A., Grey wolf optimizer. *Advances in engineering software* **2014,** 69, 46-61.
133. Mirjalili, S.; Lewis, A., The whale optimization algorithm. *Advances in engineering software* **2016,** 95, 51-67.
134. Karaboga, D. *An idea based on honey bee swarm for numerical optimization*; Citeseer: 2005.
135. Colorni, A.; Dorigo, M.; Maniezzo, V. In *Distributed optimization by ant colonies*, Proceedings of the first European conference on artificial life, 1992; Cambridge, MA: pp 134-142.
136. Wang, G.-G.; Deb, S.; Coelho, L. d. S. In *Elephant herding optimization*, 2015 3rd International Symposium on Computational and Business Intelligence (ISCBI), 2015; IEEE: pp 1-5.

137. Holland, J. H., Adaptation in natural and artificial systems: an introductory analysis with applications to biology, control, and artificial intelligence. University of Michigan press Ann Arbor: 1975.
138. Atashpaz-Gargari, E.; Lucas, C. In *Imperialist competitive algorithm: an algorithm for optimization inspired by imperialistic competition*, 2007 IEEE congress on evolutionary computation, 2007; IEEE: pp 4661-4667.
139. Kennedy, J.; Eberhart, R. In *Particle swarm optimization*, Proceedings of ICNN'95 - International Conference on Neural Networks, 27 Nov.-1 Dec. 1995, 1995; pp 1942-1948 vol.4.
140. Bayraktar, Z.; Komurcu, M.; Werner, D. H. In *Wind Driven Optimization (WDO): A novel nature-inspired optimization algorithm and its application to electromagnetics*, 2010 IEEE Antennas and Propagation Society International Symposium, 2010; IEEE: pp 1-4.
141. Deng, H.; Fannon, D.; Eckelman, M. J., Predictive modeling for US commercial building energy use: A comparison of existing statistical and machine learning algorithms using CBECS microdata. *Energy and Buildings* **2018,** 163, 34-43.